# Advancing 6D Pose Estimation in Augmented Reality - Overcoming Projection Ambiguity with Uncontrolled Imagery

Mayura Manawadu, Sieun Park, Soon-Yong Park

School of Electronic and Electrical Engineering, Kyungpook National University

mayuramanawadu@knu.ac.kr, swanyh@naver.com, sypark@knu.ac.kr

## Abstract

This study addresses the challenge of accurate 6D pose estimation in Augmented Reality (AR), a critical component for seamlessly integrating virtual objects into real-world environments. Our research primarily addresses the difficulty of estimating 6D poses from uncontrolled RGB images, a common scenario in AR applications, which lacks metadata such as focal length. We propose a novel approach that strategically decomposes the estimation of z-axis translation and focal length, leveraging the neural-render and compare strategy inherent in the FocalPose architecture. This methodology not only streamlines the 6D pose estimation process but also significantly enhances the accuracy of 3D object overlaying in AR settings. Our experimental results demonstrate a marked improvement in 6D pose estimation accuracy, with promising applications in manufacturing and robotics. Here, the precise overlay of AR visualizations and the advancement of robotic vision systems stand to benefit substantially from our findings.

## 1. Introduction

Augmented Reality (AR) has rapidly emerged as a pivotal technology of Computer Vision in the seamless blending of digital and physical worlds. The effectiveness of AR in enhancing user engagement and operational efficiency significantly relies on the accuracy of 6D pose estimation – the technology crucial for precisely aligning virtual objects within real-world settings. This accuracy is vital for creating truly immersive AR experiences. As AR extends its reach from media and entertainment to industrial and medical applications, advances in 6D pose estimation become increasingly important.

However, the rise of AR applications using uncontrolled, 'in the wild' imagery presents new challenges. These images often lack crucial metadata like focal length and are marked by variable camera parameters and diverse environmental conditions, posing significant challenges to traditional pose estimation methods. This limitation is a major obstacle in the wider adoption and practicality of AR technologies. Also, Environmental variables such as inconsistent lighting, varied background textures, and changing object orientations significantly impact the performance of pose estimation algorithms. Even though recent research works have evolved to address this challenge, they rely on non-differentiable optimizers [1] or approximations in their methods [2], limiting end-to-end trainability and accuracy. Additionally, the estimation of internal camera parameters (like focal length) alongside external pose parameters becomes increasingly complex in these dynamic settings. To tackle these challenges, our method, drawing inspiration from the neural render-and-compare strategy of Focalpose, effectively decomposes the estimation of z-axis translation from focal length. This refinement not only simplifies the estimation process but also ensures robustness and enhanced accuracy in diverse real-world applications. This advancement in pose estimation technology promises substantial improvements in AR applications, from enhancing industrial assembly precision to advancing medical visualization techniques. By elevating the accuracy and reliability of pose estimation in uncontrolled imagery, our work contributes significantly to the evolution and efficacy of AR experiences.

## 2. Related Works

In the domain of computer vision, 6-D pose estimation is crucial, particularly for augmented reality (AR) applications, also extending to robotics and Simultaneous Localization and Mapping (SLAM). Precise pose estimation is key for integrating virtual elements into real-world environments, enhancing user interaction in AR. The field has evolved from classical feature-based techniques like Fast Hierarchical Template matching and local descriptors like SIFT [3] , SURF [4] to advanced deep learning methods, notably Convolutional Neural Networks (CNNs), for direct pose regression from RGB images. This evolution includes hybrid approaches, combining traditional techniques with deep learning for refined pose accuracy. Notable developments like Deepfocal [5] for focal length estimation and CosyPose's [6] multi-view matching system have significantly influenced the field. Ponimatkin et. Al. presents FocalPose [7], an extension of CosyPose integrating focal length in pose estimation, further representing this progress. The integration of cascaded refinement transformers and approaches for reflective materials from ContourPose [8] by Zaixing et. al. suggests possibilities for more comprehensive solutions. Our research builds on FocalPose, aiming to adapt these techniques for dynamic AR environments. We propose a method that decomposes translation along z axis in 6D pose from focal length, employing a render-and-compare approach. From this work, we aim to overcome existing limitations in AR applications on uncontrolled imagery.

## 3. Methodology

In our work, we present a methodology based on Focalpose, by refining the accuracy of joint 6D pose and camera focal length estimation from single RGB images. Our improvements focus on optimizing the neural-render and compare strategy, by decomposing the simultaneous estimation of focal length and z-axis translation.

### 3.1 Render and Compare Methodology

The "render and compare" method for pose estimation is an approach that involves generating synthetic images of an object in various poses and then comparing these renders to real-world images. This technique leverages deep learning algorithms to refine pose predictions, ensuring accuracy by minimizing the difference between rendered synthetic and real images. Recently, research works such as CosyPose [6], DeepIM [9], Focalpose [7] have shown promising results using refined networks. making it more viable for real-time applications in robotics, augmented reality, and other fields.

### 3.2 6D Pose Estimation of "In-The-Wild" images.

"In the wild" images in computer vision refer to photos or visual data captured in natural, uncontrolled environments, as opposed to those taken in controlled settings like studios. In those images, the focal length of the camera used to capture the image is often unknown or not provided. Focal length is a crucial parameter in traditional photography and computer vision as it influences the field of view and the perceived depth and scale of objects in an image. While majority of the research works have been focused on pose estimation using single RGB images based on controlled imagery, only few research works like Focalpose have been developed based on uncontrolled imagery [1], [2].

### 3.3 Proposed Approach

In our work, we address an inherent ambiguity of Focalpose where simultaneous predictions for Z-axis translation and focal length can overlap, leading to uncertain results. This issue arises from estimating both parameters at once, creating a situation analogous to finding variables $A$ and $B$ in the equation $C = AB$, where $C$ is constant, allowing for multiple valid solutions. By fixing the Z-axis translation, we significantly reduce this complexity, achieving more precise and distinct outcomes. In the context of AR, this precision is critical for accurately placing virtual objects in real-world scenes. By stabilizing Z-axis translation while accurately estimating pose and focal length, we improve the depth and scale estimation of objects, ensuring virtual objects are rendered at correct sizes and positions. Additionally, a precise understanding of the camera's focal length is vital for calibrating AR systems effectively. Proper calibration is key to aligning virtual and real-world coordinates, essential for a seamless AR experience. An incorrect focal length can result in mismatches in object sizes and distances, disrupting the AR illusion. While fixing the Z-axis translation introduces the challenge of determining it, in AR, techniques like ray casting can be employed. Using ray casting, a ray from the camera identifies depth by intersecting with scene objects, enabling precise placement of virtual objects at consistent real-world depths. This stabilizes Z-axis translation and improves pose and focal length estimates, ensuring accurate rendering of virtual objects in size and position.

### 3.4 Developed Architecture

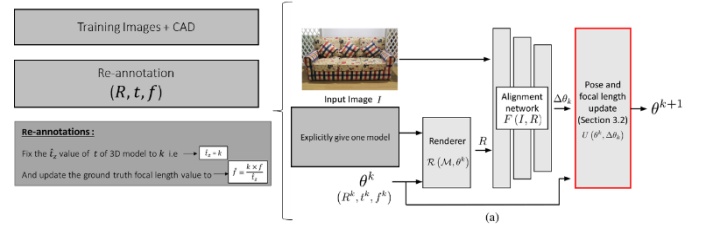

Fig. 1. Enhanced Focalpose Architecture

The modified architecture from our work is represented in Fig. 1. Here, to set the translation along the z axis ($t_z$) to an arbitrary constant, we re-annotated the ground truth annotations and following that, we rescaled focal length as shown in the following equations.

$$t_{z_{new}} = k \quad (1)$$

$$f_{new} = f_{old} \frac{k}{t_{z_{old}}} \quad (2)$$

The re-annotated ground truth data are input into the neural network, which was trained using customized update rules and loss functions. The network receives an RGB image and its corresponding 3D model as inputs. Using the Mask-RCNN network, the object in the RGB image is identified, and the 3D model is rendered with an initial estimate of pose and focal length, denoted as $\theta^k$, using renderer $R$. The alignment network $F$, built on the ResNet-50 architecture, calculates the necessary update of $\Delta\theta^k$ through the render-and-compare method, utilizing the RGB image and 3D model. The subsequent pose and focal length are determined by a specially designed non-linear update rule $U$. In our methodology, the loss function from Focalpose is adapted, incorporating the fixed z-axis translation and adjusted focal length. The update rule $U$ is specifically modified for our approach as follows.

$$x^{k+1} = \frac{v_x^k}{f^{k+1}} k + x^k \quad (3)$$

$$y^{k+1} = \frac{v_y^k}{f^{k+1}} k + y^k \quad (4)$$

Here $v_x^k$ & $v_y^k$ are the components of $\Delta\theta^k$ which corresponds to the update of $x$ and $y$ axis translations predcited by $F$. The update rule for focal length is given below, where it uses the scaled focal length of Equation (2).

$$f^{k+1} = (e^{v_f^k}) f_{old} \frac{k}{t_z} \quad (5)$$

## 4. Experiments

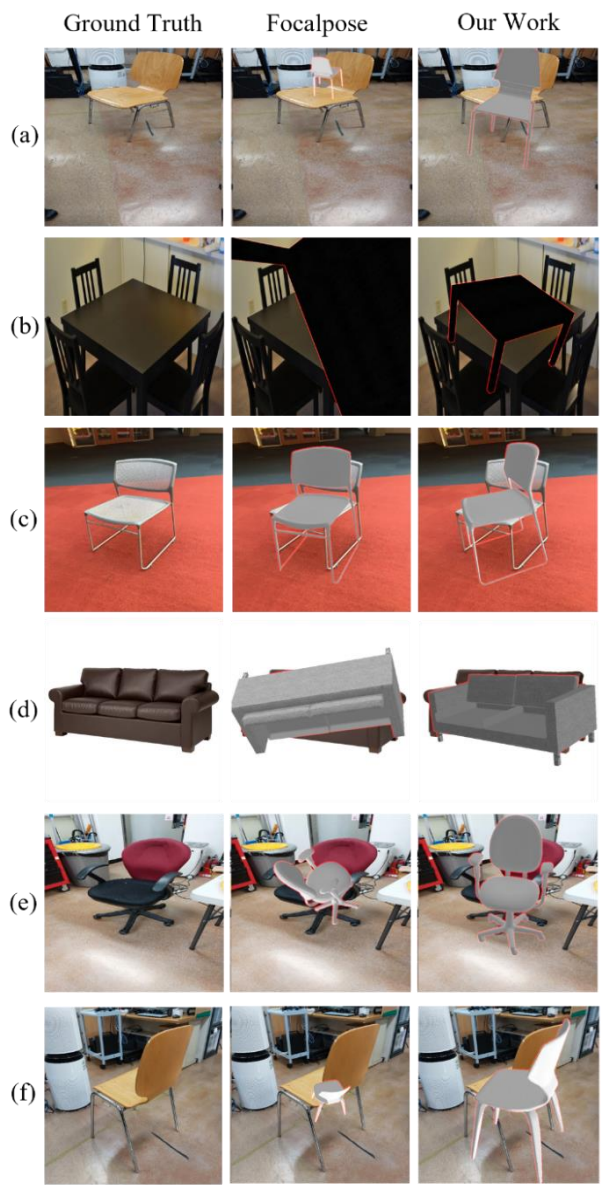

Fig. 2. Comparison of Focalpose with Ours

Upon decomposing the Z-axis translation, we conducted a qualitative analysis on the Pix3D Dataset, comparing it against the Focalpose method as given in Fig. 2. The enhanced accuracy of our approach in object overlaying, underscoring the limitations inherent in Focalpose, which often produces ambiguous results. Our method, with its separate estimation of focal length, yields more precise outcomes. For an example, Figure 1(a) illustrates the challenges in simultaneous $t_z$ and focal length prediction. Notably, as seen in Figure 2(b), while both methods occasionally fail in precise pose estimation, the focal length accuracy of our approach remains more accurate. That result is reasonable to the fact that both methods were trained without the integration of synthetic data, due to the limitations in computational resource. Nevertheless, the precise estimations achieved by our method on real datasets suggest that its efficacy could be further improved through training on a dataset with a higher proportion of synthesized data which is less noisy when compare to real dataset of Pix3D.

## 5. Conclusion

In the field of 6D pose estimation, vital for augmented reality (AR) and robotics, our research presents a strategy by decomposing camera-object 6D pose and camera focal length estimation from RGB images, simplifying complexities arising from their interdependence. By fixing variables like Z-axis translation, our approach reduces computational complexity, crucial for AR on mobile devices with limited processing power, enabling real-time performance without significant accuracy loss. This adaptability enhances robustness in diverse environments, ensuring intuitive and natural user interactions in AR, as the system accurately renders objects at correct distances, thereby improving user experience. This methodology, tested on real-world data, promises a foundation for more precise and reliable pose estimation techniques.

## Acknowledgements

This paper was conducted with the support of the Information and Communication Planning Evaluation Institute with funding from the government (Ministry of Science and ICT) in 2021. (No.2021-0-00320; real-time target XR generation and transformation/enhancing technology development).